\documentclass[letterpaper]{article} 
\usepackage{aaai2027} 
\usepackage[hyphens]{url} 
\usepackage{graphicx} 
\usepackage{natbib} 
\usepackage{caption} 
\usepackage{amsmath}
\usepackage{amssymb}
\usepackage{booktabs}
\usepackage{multirow}
\usepackage{colortbl}
\usepackage{xcolor}
\usepackage{algorithm}
\usepackage{listings}
\definecolor{refgray}{gray}{0.40}
\newcommand{\grouphead}[1]{%
  \multicolumn{5}{@{}>{\columncolor{black!10}[0pt][0pt]}l@{}}{\textit{\textbf{#1}}}}
\newcommand{\dimrow}[5]{%
  \textcolor{refgray}{#1} & \textcolor{refgray}{#2} & \textcolor{refgray}{#3} &
  \textcolor{refgray}{#4} & \textcolor{refgray}{#5}}
\definecolor{rowgray}{gray}{0.92}
\definecolor{fadetext}{gray}{0.5}

\lstdefinestyle{pythonstyle}{
    language=Python,
    basicstyle=\footnotesize\ttfamily,
    numbers=none,
    aboveskip=0em,
    belowskip=0em,
    showstringspaces=false,
    tabsize=4,
    breaklines=true,
    moredelim=[is][\bfseries]{@}{@} 
}

\title{Rethinking Pixel Mean Flows via Interval Denoiser}
\author{
    Alexander Zaytsev\textsuperscript{1},
    Dmitry Baranchuk\textsuperscript{2},
    Alexander Korotin\textsuperscript{3,4},
    Aibek Alanov\textsuperscript{1,4,5}
}
\affiliations{
    \textsuperscript{1}HSE University\\
    \textsuperscript{2}Yandex Research\\
    \textsuperscript{3}Applied AI Institute\\
    \textsuperscript{4}AXXX\\
    \textsuperscript{5}FusionBrain Lab
}

\begin{document}
\nocopyright
\maketitle

\begin{abstract}

Modern diffusion and flow-based models are increasingly moving toward few-step, latent-free generation to bypass the computational overhead of multi-step sampling and the reconstruction bottlenecks of external autoencoders. We propose the \textit{Interval Denoiser}, a theoretically rigorous framework for latent-free generation. Derived directly from the flow matching ODE, it establishes an exact analytical mapping for intermediate trajectory states. Unlike prior formulations, our prediction is shown to reside on a low-dimensional manifold across any time interval, making the regression tractable for a network operating directly on pixels. Furthermore, by avoiding empirical algebraic substitutions, our formulation correctly isolates the pure time derivative to prevent biased gradient evaluations and ensure exact first-order optimization. By analyzing this objective, we equip our framework with residual clipping and a time-sampling curriculum, enabling effective long-interval training and improving few-step performance. Trained from scratch on ImageNet $256{\times}256$, our model achieves an FID of 4.55 in one step (1-NFE) and 3.98 in two steps (2-NFE) without perceptual losses.

\end{abstract}
\section{Introduction}
\begin{figure}[tb]
    \centering
    \includegraphics[width=0.97\columnwidth]{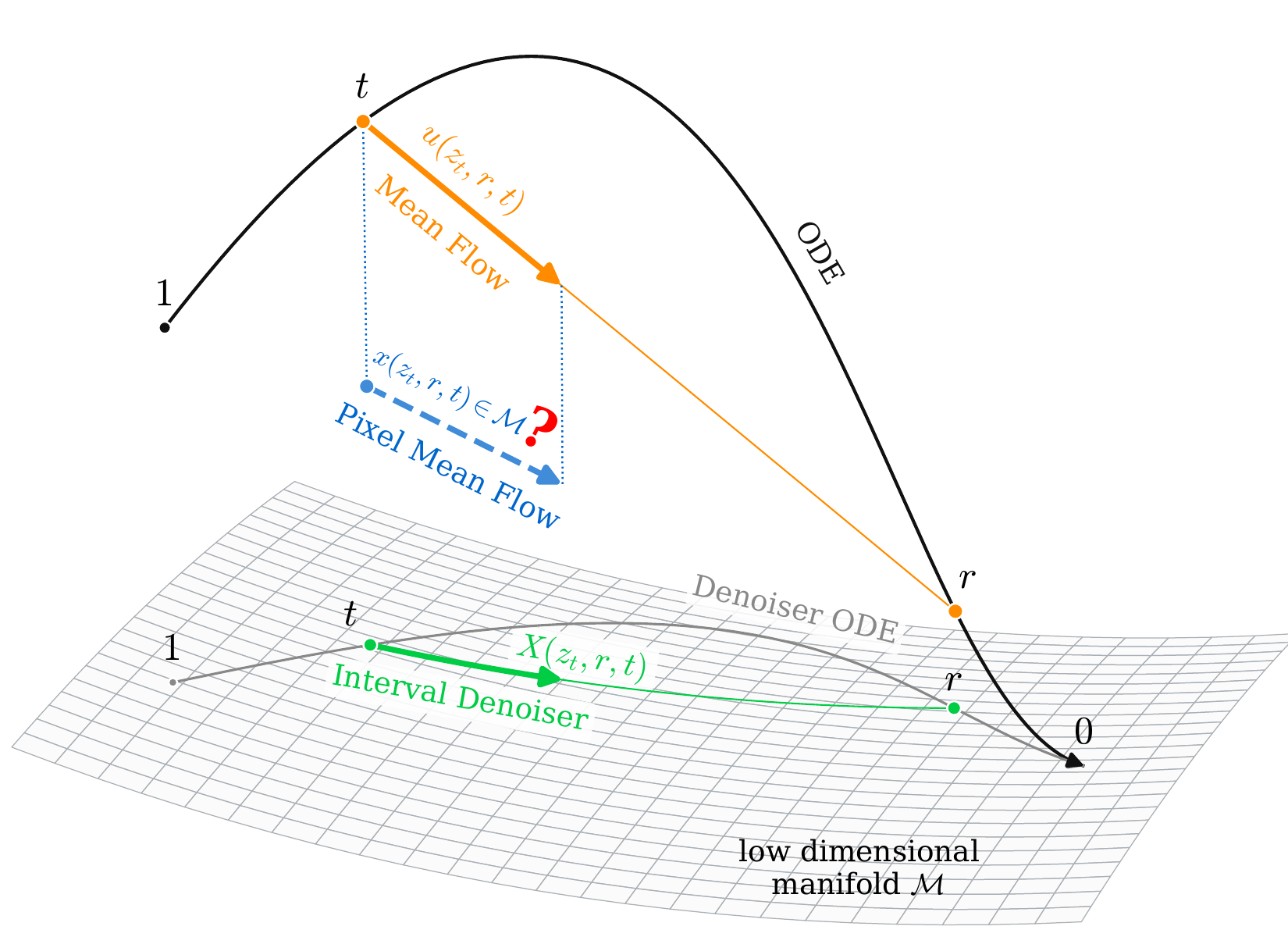}
    \caption{Geometry of prediction targets. MeanFlow (orange) targets the average velocity $u(z_t, r, t)$ in the high-dimensional ambient space. Pixel MeanFlow (blue) uses an empirical algebraic substitution that has not been shown to reside on the low-dimensional manifold $\mathcal{M}$. In contrast, our Interval Denoiser $X(z_t, r, t)$ (green) establishes an exact analytical mapping that resides on the denoised image manifold across any time interval.}
    \label{fig::x_vs_y}
\end{figure}
Diffusion models and their flow-based variants \cite{ho2020ddpm, song2021score, lipman2023flowmatching} are highly effective generative frameworks that simulate continuous-time ordinary differential equations (ODEs). However, they require multi-step numerical integration and rely on compressed latent spaces \cite{rombach2022ldm} to manage high-dimensional data, which limits pixel-level fidelity. Recently, frameworks like Consistency Models \cite{song2023consistency}, Consistency Trajectory Models (CTM) \cite{kim2024ctm}, and MeanFlow \cite{geng2025meanflow, geng2026improvedmeanflows} have drastically reduced sampling steps, while other works have demonstrated the feasibility of operating directly in raw pixel space \cite{li2025backtobasics, lei2026epg}. Together, these parallel advances pave the way for few-step, latent-free generative modeling.

Despite progress by methods like Pixel MeanFlow (pMF) \cite{lu2026pixelmeanflow} in the latent-free regime, their fundamental capabilities remain limited. To achieve image-space predictions, pMF relies on an empirical algebraic substitution applied to the Improved MeanFlow \cite{geng2026improvedmeanflows} objective. Without a formal derivation, this substitution lacks a solid mathematical foundation. Furthermore, inserting this substitution into the loss reveals the exact objective being minimized: when written in terms of the image prediction network, the loss acquires extra spatial prediction terms trapped inside the stop-gradient operator alongside the JVP, causing biased parameter updates.

\paragraph{Contributions.} In this work, we propose a principled and theoretically rigorous framework for few-step latent-free generation. We analyze the flow matching ODE and derive an explicit mapping for intermediate trajectory states termed the \textit{Interval Denoiser}. This formulation projects the generation trajectory directly onto the well-structured, low-dimensional manifold of denoised images \cite{vincent2008extracting, li2025backtobasics} (see Fig.~\ref{fig::x_vs_y}), providing a highly tractable and mathematically sound regression for the network.

Building upon this foundation, our exact formulation formally derives the empirical algebraic substitution used in pMF and naturally recovers the decoder parameterization of CTM. To further improve generation over large integration intervals, we incorporate two critical training strategies. First, through an analysis of the regression target, we demonstrate why \textit{residual clipping} \cite{lu2025scm, peng2026facm} is necessary to prevent severe signal suppression in this regime. Second, we apply a \textit{distributional curriculum} for time sampling \cite{curriculum2024flowmatching}, showing that it allows shifting the training focus from short to large intervals.

Evaluated on the ImageNet $256 \times 256$ benchmark \cite{deng2009imagenet}, our model, trained entirely from scratch in pixel space, achieves an FID of 4.55 with a single function evaluation (1-NFE), which improves to 3.98 with two steps (2-NFE). Furthermore, because perceptual losses artificially lower evaluation metrics {\cite{kynkaanniemi2023role,song2024improvedct}}, we omit them and isolate our comparison to models that do not rely on such losses, achieving state-of-the-art generation quality in 1-NFE and 2-NFE.

\section{Related Work}

\paragraph{Direct Pixel-Space Generation.}
Diffusion \cite{ho2020ddpm, song2021score} and flow matching \cite{lipman2023flowmatching, albergo2025stochastic} typically operate in the compressed latent spaces of pre-trained autoencoders \cite{rombach2022ldm}. While latent representations reduce computational overhead, they introduce reconstruction bottlenecks that limit fine-grained fidelity. Operating directly in pixel space provides a latent-free alternative, yet it exposes the network to high-dimensional inputs. Recent work has observed that Vision Transformer (ViT) \cite{dosovitskiy2021an} architectures degrade rapidly when the dimensionality per token becomes too large \cite{chen2025deconstructing, yao2025reconstruction, shi2025latent}. Furthermore, while images naturally reside on a structured, low-dimensional manifold \cite{chapelle2006semi, vincent2008extracting}, predicting unstructured high-dimensional noise or velocity fields in pixel space is difficult \cite{li2025backtobasics}. To address this, recent methods decouple the prediction and loss spaces \cite{karras2022elucidating, lu2026pixelmeanflow}. By tasking the network to output a denoised image ($x$-prediction), the prediction remains anchored to the tractable data manifold, which can then be algebraically transformed to optimize standard velocity objectives \cite{lu2026pixelmeanflow}.

\paragraph{Few-Step Generative Models.} 
To bypass the numerous NFEs required by numerical ODE solvers, various fast-forward frameworks have been developed. Consistency Models \cite{song2023consistency} and CTM \cite{kim2024ctm} learn mapping functions that enable large discrete transitions along the generation trajectory. Alternatively, the MeanFlow family \cite{geng2025meanflow, geng2026improvedmeanflows} achieves few-step sampling by predicting the average velocity over a discrete time interval. Recently, pMF \cite{lu2026pixelmeanflow} adapted these concepts for latent-free generation by combining the MeanFlow objective with an $x$-space prediction. However, pMF introduces an empirical substitution without a formal ODE parameterization. Furthermore, it uses auxiliary perceptual losses \cite{zhang2018lpips}, which can artificially improve scores on standard metrics by shifting the optimization target away from true distribution matching.

\section{Background}

\paragraph{Flow Matching.}
Flow matching \cite{lipman2023flowmatching} learns a vector field to transport a standard Gaussian prior to a data distribution. For clean data $x_0 \sim p_{\text{data}}$ at $t=0$ and noise $\epsilon \sim \mathcal{N}(0, I)$ at $t=1$, the linear probability flow path and conditional velocity $v_t$ are:
\begin{equation}
    z_t = (1 - t)x_0 + t\epsilon, \quad v_t = \epsilon - x_0.
\end{equation}
Because the marginal velocity $v(z_t, t) = \mathbb{E}[v_t \mid z_t]$ is intractable, models approximate it with a neural network $v_\theta(z_t, t)$ by minimizing the regression loss against $v_t$:
\begin{equation}
    \mathcal{L}_{\text{FM}} = \mathbb{E}_{t, x_0, \epsilon} \big[ \| v_\theta(z_t, t) - v_t \|_2^2 \big].
\end{equation}
Samples are generated by solving the ODE $dz_t/dt = v_\theta(z_t, t)$ from $t=1$ to $t=0$.

\paragraph{The MeanFlow Family.}
To enable large generation steps, MeanFlow \cite{geng2025meanflow} predicts the average velocity over a discrete interval $[r, t]$:
\begin{equation}
    u(z_t, r, t) = \frac{1}{t - r} \int_{r}^{t} v(z_\tau, \tau) d\tau.
\end{equation}
Approximating this target with a network $u_\theta(z_t, r, t)$ yields the sampling step:
\begin{equation}
    z_r = z_t - (t - r)u_\theta(z_t, r, t).
\end{equation}
To establish a network-independent regression target matching $v_t$, Improved MeanFlow \cite{geng2026improvedmeanflows} computes the time derivative via a stop-gradient Jacobian Vector Product ($\text{JVP}_{\text{sg}}$) along the network's instantaneous velocity $v_\theta$. Omitting inputs for brevity, the objective is:
\begin{equation}
    \mathcal{L}_{\text{iMF}} = \mathbb{E}_{t, r, x_0, \epsilon} \big[ \| u_\theta + (t - r)\text{JVP}_{\text{sg}}(u_\theta; v_\theta) - v_t \|_2^2 \big].
\end{equation}
For pixel-space generation, pMF \cite{lu2026pixelmeanflow} introduces an image-space network $X_\theta(z_t, r, t)$ via an empirical algebraic substitution:
\begin{equation} 
    u_\theta(z_t, r, t) = \frac{z_t - X_\theta(z_t, r, t)}{t}.
\end{equation}
However, expressing the loss via the image network $X_\theta$ under this substitution reveals that the objective accumulates extra spatial prediction terms trapped inside the stop-gradient alongside the JVP, causing biased updates. Therefore, pMF has two key limitations: it uses an empirical substitution without an ODE derivation, and it offers no proof that predictions remain on the low-dimensional manifold.

\paragraph{Consistency Trajectory Models.}
CTM \cite{kim2024ctm} learns any-to-any timestep transitions. To enforce the boundary condition $f_\theta(z_t, t, t) = z_t$, CTM isolates a data predictor $g_\theta(z_t, t, r)$ to define its transition mapping:
\begin{equation}
\label{eq:ctm_mapping}
    f_\theta(z_t, t, r) = \frac{r}{t}z_t + \frac{t - r}{t}g_\theta(z_t, t, r).
\end{equation}
CTM optimizes this mapping via trajectory consistency, enforcing that a direct step from $t$ to $r$ aligns with an intermediate step at $s \in (r, t)$ evaluated by a stop-gradient target network $f_{\text{target}}$:
\begin{equation}
    \mathcal{L}_{\text{CTM}} = \mathbb{E}_{t, s, r, x_0, \epsilon} \big[ \| f_\theta(z_t, t, r) - \text{sg}(f_{\text{target}}(z_s, s, r)) \|_2^2 \big].
\end{equation}
We demonstrate that the exact structural form of $f_\theta$ naturally emerges within our framework.

\section{Interval Denoiser Models}

We introduce the pixel Interval Denoiser (pID) for few-step latent-free generation. Derived from the flow matching ODE, its predictions reside on the low-dimensional manifold, making generation tractable in pixel space.

\subsection{Fundamentals of Interval Denoising}

To achieve this, we seek an update function for the step from $z_t$ to $z_r$ that depends strictly on the instantaneous denoiser. This completely isolates the network's output from noise.

\paragraph{Analyzing the Flow Matching ODE.}
For flow matching \cite{lipman2023flowmatching, albergo2025stochastic} operating in image space, the ODE is commonly defined using the instantaneous denoiser $x(z_t, t)$ at time $t$:
\begin{equation} 
\label{eq::x_fm} 
\frac{d z_t}{dt} = \frac{z_t - x(z_t, t)}{t}. 
\end{equation} 
To derive this update function, we divide both sides of Eq.~\ref{eq::x_fm} by $t$ and rearrange the terms to form an exact differential:
\begin{equation}
\label{integrating_term}
\frac{1}{t}\frac{d z_t}{dt} -\frac{z_t}{t^2} = -\frac{x(z_t, t)}{t^2} \quad \Rightarrow \quad \frac{d}{dt}\left(\frac{z_t}{t}\right)= - \frac{x(z_t, t)}{t^2}.
\end{equation} 
Integrating from a target time $r$ to the current time $t$ provides the exact state update:
\begin{equation}
\label{for_sampling}    
z_r = \frac{r}{t}z_t + r \int^t_r \frac{x(z_\tau, \tau)}{\tau^2} d\tau. 
\end{equation} 
This confirms that the step from $z_t$ to $z_r$ relies entirely on the integral of the scaled instantaneous denoiser, without requiring velocity or noise.

\paragraph{Definition of the Interval Denoiser.}
To parameterize this integral, we define the \textit{Interval Denoiser} $X(z_t, r, t)$ as a normalized, weighted aggregation of instantaneous predictions over the time interval $[r, t]$:
\begin{equation}
\label{proj}
X (z_t, r, t) = \frac{t \cdot r}{t - r} \int^t_r \frac{x(z_\tau, \tau)}{\tau^2} d\tau. 
\end{equation}

\paragraph{The Generalized Manifold Hypothesis.}
We establish that Eq.~\ref{proj} constitutes a valid mathematical expectation.

\noindent \textbf{Proposition 1.} \textit{Assume the denoiser is optimal, such that ${x(z_\tau, \tau) = \mathbb{E}[x_0 \mid z_\tau]}$. Then for $t > r$, ${X(z_t, r, t) = \mathbb{E}_{\tau}[\mathbb{E}[x_0 \mid z_\tau]]}$, and at $r = t$ it holds that ${X(z_t, t, t) = x(z_t, t)}$.}

\noindent \textbf{Proof.} For $t > r$, the weighting term $p(\tau) = \frac{t \cdot r}{t - r} \tau^{-2} \ge 0$ integrates exactly to $1$ on $[r, t]$, serving as a valid probability density function. As $r \to t$, the mean value theorem for definite integrals yields $\lim_{r \to t} X(z_t, r, t) = x(z_t, t)$. $\blacksquare$

By Prop.~1, $X$ averages denoiser outputs along a single trajectory, all
estimating the same clean image $x_0$. For any interval, the average is again
an estimate of $x_0$, so it lies in the same low-dimensional set of denoised
images. The generalized manifold hypothesis \cite{lu2026pixelmeanflow}
therefore holds at every $(r,t)$. As established by \citet{li2025backtobasics}, predicting this on-manifold target makes direct pixel-space learning tractable, because the model can focus on learning the underlying data manifold instead of preserving high-dimensional noise or velocity vectors across ambient space.

\paragraph{Interval Denoiser Identity.}
To construct a training objective, we isolate the integral in Eq.~\ref{proj}:
\begin{equation}
\label{diff} 
\frac{t - r}{t \cdot r}  X (z_t, r, t) = \int^t_r \frac{x(z_\tau, \tau)}{\tau^2} d\tau. 
\end{equation} 
Differentiating both sides with respect to $t$ and multiplying by $t^2$ yields the fundamental \textit{Interval Denoiser Identity}:
\begin{equation}
\label{identity}
 X (z_t, r, t)  +  \frac{t(t-r)}{r} \frac{d}{dt}X(z_t, r, t) = x(z_t, t). 
\end{equation}
\subsection{Training and Inference Strategy}

We now turn the Interval Denoiser identity into a practical training objective. Specific architectural details are provided in Appendix~\ref{sec:impl}.

\paragraph{Computing the Time Derivative.}
Evaluating Eq.~\ref{identity} requires computing the total time derivative $\frac{d}{dt}X_\theta(z_t, r, t)$.
\begin{equation}
    \frac{d}{dt}X_\theta(z_t, r, t) = \partial_z X_\theta \frac{dz_t}{dt} + \partial_r X_\theta \frac{dr}{dt} + \partial_t X_\theta \frac{dt}{dt}.
\end{equation}
Since the target time $r$ is independent of $t$, we have $\frac{dr}{dt}=0$, and $\frac{dt}{dt}=1$. Following \citet{geng2026improvedmeanflows}, we evaluate the trajectory state update $\frac{dz_t}{dt}$ using the network output $x_\theta(z_t, t)$, which targets the boundary value $X(z_t, t, t) = x(z_t, t)$ of Prop.~1. Substituting this yields:
\begin{equation}
\label{der}
\frac{d}{dt}X_\theta(z_t, r, t) = \partial_z X_\theta \left(\frac{z_t - x_\theta(z_t, t)}{t}\right) + \partial_t X_\theta.
\end{equation}
This is efficiently computed via a JVP along the tangent vector $\big[\frac{z_t - x_\theta}{t}, 0, 1\big]$. To avoid division by $t$, we absorb $t$ from the coefficient $\frac{t(t-r)}{r}$ directly into the JVP. This scales the tangent vector to $[z_t - x_\theta, 0, t]$, yielding a tractable expression denoted $\text{JVP}_{X_\theta}$.

\paragraph{Training.}
Because the exact denoiser $x(z_t, t)$ is intractable, we substitute the ground-truth image $x_0$. To avoid higher-order derivatives, the JVP uses a stop-gradient network copy $\theta^-$, yielding the regression objective:
\begin{equation}
\mathcal{L}(\theta) = \mathbb{E}_{t, r, x_0, \epsilon} \Big[ \big\| X_\theta(z_t, r, t) + \frac{t - r}{r} \text{JVP}_{X_{\theta^-}} - x_0  \big\|_2^2 \Big].
\end{equation}
The training procedure is summarized in Alg.~\ref{alg:training}.

\paragraph{Sampling.}
At inference, substituting the network $X_\theta$ back into the exact update rule (Eq.~\ref{for_sampling}) yields:
\begin{equation}
\label{sampling}
z_r = \frac{1}{t} \Big(r z_t + (t-r)X_\theta(z_t, r, t) \Big). 
\end{equation}
The sampling procedure is summarized in Alg.~\ref{alg:sampling}.

\begin{algorithm}[tb]
\caption{Interval Denoiser: Training.}
\begin{lstlisting}[style=pythonstyle]
# net: online interval denoiser network
# x0: clean images in pixels

t, r = sample_t_r()
e = randn_like(x0)
z = (1 - t) * x0 + t * e

x = net(z, t, t)
X, dXdt = @jvp@(net, (z, r, t), (z - x, 0, t))

error = X + (t - r)/r * @stopgrad@(dXdt) - x0
loss = @metric@(error)
\end{lstlisting}
\label{alg:training}
\end{algorithm}

\begin{algorithm}[tb]
\caption{Interval Denoiser: Sampling.}
\begin{lstlisting}[style=pythonstyle]
# net: trained interval denoiser network
# timesteps: array from t_0=1 to t_N=0
# shape: desired image dimensions

z = randn_like(shape)
for i in range(N):
    t, r = timesteps[i], timesteps[i+1]
    X_pred = net(z, r, t)
    
    z = (r * z + (t - r) * X_pred) / t
return z
\end{lstlisting}
\label{alg:sampling}
\end{algorithm}

\subsection{Relation to Prior Work}

Our framework formally connects to few-step models. First, the Interval Denoiser relates to MeanFlow as the instantaneous denoiser relates to velocity, deriving pMF's empirical substitution. Second, it explains CTM: its preconditioned mapping matches our sampling update, and the continuous-time limit of its trajectory loss recovers our formulation.

\paragraph{Connection to MeanFlow.}
\label{sec:connection_meanflow}

MeanFlow updates the trajectory from $t$ to $r$ via average velocity $u(z_t, r, t)$: $z_r = z_t - (t-r)u(z_t, r, t)$. Rearranging our sampling update (Eq.~\ref{sampling}) yields an identical form:
\begin{equation}
\label{eq:update_ours}
    z_r = \frac{r \cdot z_t + (t-r)X(z_t, r, t)}{t} = z_t - (t-r) \frac{z_t - X(z_t, r, t)}{t}.
\end{equation}
Equating these two trajectory update steps formally connects the Interval Denoiser to average velocity:
\begin{equation}
\label{eq:relationship}    
    u(z_t, r, t) = \frac{z_t - X(z_t, r, t)}{t}. 
\end{equation} 
Whereas pMF introduces this mapping as an empirical substitution, our derivation proves it is a direct mathematical consequence of predicting in image space, mirroring the standard denoiser-velocity relationship.

\paragraph{Biased Optimization via Algebraic Substitutions.}
pMF constructs its loss by substituting $u_\theta = (z_t - X_\theta)/t$ into the Improved MeanFlow objective. This forces the total time derivative to expand. Denoting $X_\theta \equiv X_\theta(z_t, r, t)$ and the boundary $x_\theta \equiv X_\theta(z_t, t, t)$, the resulting pMF training objective takes the following form:
\begin{equation}
    \mathcal{L}_{\text{pMF}} = \frac{1}{t^2}\mathbb{E} \Big\| X_\theta + (t-r)\text{sg}\Big( \frac{d X_\theta}{dt} - \frac{X_\theta - x_\theta}{t}  \Big) - x_0 \Big\|_2^2.
\end{equation}
Similarly, substituting this parameterization into the original MeanFlow objective also yields an additional spatial term trapped inside the stop-gradient:
\begin{equation}
    \mathcal{L}_{\text{MF}} = \frac{1}{t^2} \mathbb{E} \left[  \left\| X_\theta + (t - r) \text{sg}\left(\frac{dX_\theta}{dt} - \frac{X_\theta}{t} \right) - \frac{r}{t} x_0 \right\|_2^2 \right].
\end{equation}

Both formulations trap spatial predictions $X_\theta$ or $x_\theta$ inside the stop-gradient $\text{sg}(\cdot)$. Masking these parameters yields a biased update diverging from the analytical gradient. Our formulation resolves this by isolating the pure time derivative. Applying the stop-gradient to the JVP hides no spatial parameters, ensuring exact first-order optimization.

\paragraph{Interval Denoiser with MeanFlow Loss.}
\label{app:beta_derivation}

In comparison to inserting the substitution directly into the empirical loss, we return to the fundamental differential identities. Evaluating the MeanFlow objective through this theoretical connection naturally induces a scaling factor. This establishes the mathematical equivalence between the two regression spaces. Dependencies on $z_t, r$, and $t$ are omitted for brevity.

\noindent \textbf{Proposition 2.} \textit{Given $u = \frac{z_t - X}{t}$, $v = \frac{z_t - x}{t}$, and $\frac{dz_t}{dt} = v$, the MeanFlow identity is equivalent to the Interval Denoiser identity scaled by $\frac{r}{t^2}$:}
\begin{equation*}
    v - u - (t-r)\frac{du}{dt} = \frac{r}{t^2} \bigg[ X + \frac{t(t-r)}{r}\frac{dX}{dt} - x \bigg]. 
\end{equation*}

\noindent \textbf{Proof.} Using the quotient rule and substituting the ODE $\frac{dz_t}{dt} = \frac{z_t - x}{t}$, the total time derivative $\frac{du}{dt}$ expands as:
\begin{equation}
\label{app:thisderivate}
    \frac{du}{dt} = \frac{1}{t}\frac{dz_t}{dt} - \frac{z_t - X}{t^2} - \frac{1}{t}\frac{dX}{dt} = \frac{X - x}{t^2} - \frac{1}{t}\frac{dX}{dt}.
\end{equation}
Substituting this derivative, alongside $u$ and $v$, into the MeanFlow residual yields:
\begin{equation}
\label{eq:pmf_residual_expansion}
\begin{split}
    &v - u - (t-r)\frac{du}{dt} \\
    &= \frac{X - x}{t} - (t-r)\left( \frac{X - x}{t^2} - \frac{1}{t}\frac{dX}{dt} \right) \\ 
    &= \frac{r}{t^2} \left[ X + \frac{t(t-r)}{r}\frac{dX}{dt} - x \right].
\end{split}
\end{equation}
Taking the squared $L_2$ norm of this residual extracts the $\frac{r^2}{t^4}$ scaling factor for the loss, establishing the formal equivalence of the two training objectives. $\blacksquare$

\begin{figure*}[t]
    \centering
    \includegraphics[width=\textwidth]{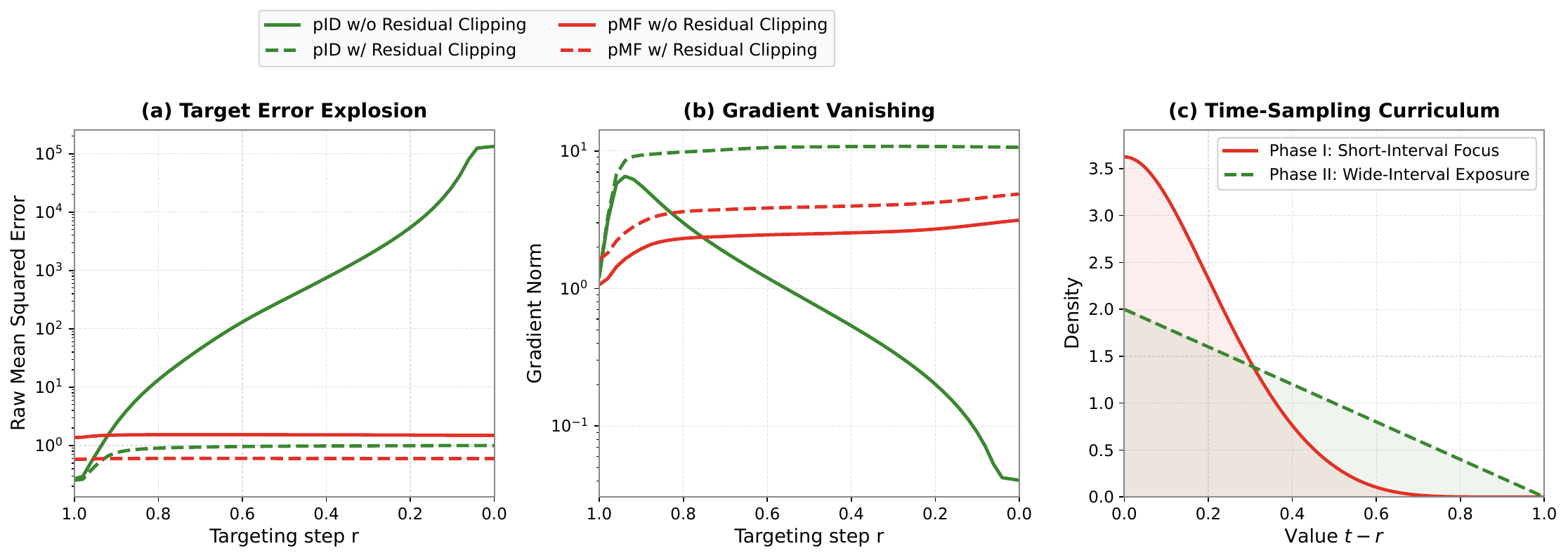} 
    \caption{Training dynamics and interval sampling analysis. We compare our pID with pMF. \textbf{(a)} Raw Mean Squared Error of the regression target at $t=1.0$, exploding for pID as $r \to 0$ without residual clipping. \textbf{(b)} Gradient norms for the logarithmic loss (Eq.~\ref{loss}) evaluated at $t=1.0$, vanishing for pID over wide intervals without clipping. While pMF is always stable, residual clipping stabilizes pID gradients. \textbf{(c)} Probability density of the integration interval $t - r$ across curriculum phases, shifting from short-interval focus (Phase I, logit-normal) to wide-interval exposure (Phase II, uniform).}
    \label{fig:residual_clipping}
\end{figure*}

\paragraph{Connection to Consistency Trajectory Models.}
\label{sec:ctm_connection}

Comparing CTM's preconditioned mapping (Eq.~\ref{eq:ctm_mapping}) with our
sampling update (Eq.~\ref{sampling}) immediately establishes the structural
equivalence $X_\theta \equiv g_\theta$.
This correspondence extends from the parameterization to the objective.

The discrete CTM objective minimizes trajectory discrepancy against a
target network across a finite step $h$:
\begin{equation}
\mathcal{L}_{\text{CTM}} = \| f_\theta(z_t, t, r) - \text{sg}(f_{\text{target}}(z_{t+h}, t+h, r)) \|_2^2.
\end{equation}
Under continuous-time teacher dynamics, where the target network converges
to the online network as $h \to 0^+$ ($f_{\text{target}} \to f_\theta$)
\cite{song2023consistency, lu2025scm}, dividing by $h^2$ and taking the
limit converts this difference into the total time derivative:
\begin{equation}
    \lim_{h \to 0^+} \frac{1}{h^2} \mathcal{L}_{\text{CTM}} = \left\| \frac{d}{dt} f_\theta(z_t, t, r) \right\|_2^2.
\end{equation}
Expanding this derivative with $X_\theta \equiv g_\theta$ yields:
\begin{equation}
    \frac{d}{dt} f_\theta = -\frac{r}{t^2}z_t + \frac{r}{t}\frac{dz_t}{dt} + \frac{r}{t^2}X_\theta + \frac{t-r}{t}\frac{dX_\theta}{dt}.
\end{equation}
Substituting $\frac{dz_t}{dt} = \frac{z_t - x}{t}$ and applying the standard
supervision substitution of $x$ by the ground-truth $x_0$ yields:
\begin{equation}
    \frac{d}{dt} f_\theta = \frac{r}{t^2} \left[ X_\theta + \frac{t(t-r)}{r}\frac{dX_\theta}{dt} - x_0  \right].
\end{equation}
This is our regression residual, scaled by $\frac{r^2}{t^4}$. CTM measures
the same quantity over a finite step rather than in the limit.

\subsection{Design Decisions}

\label{sec:design_decisions}

We adopt the following key design choices:

\paragraph{Logarithmic Objective.}
To stabilize training, we adopt a logarithmic objective. Its gradient, $\nabla_\theta \log(e + \delta) = \frac{1}{e + \delta} \nabla_\theta e$ (where $e$ is the squared error and $\delta$ is a small positive constant), exactly recovers the gradient of the adaptively weighted loss $ \frac{e}{\text{sg}(e + \delta)^p}$ with $p=1$ used in recent few-step models \cite{geng2026improvedmeanflows, peng2026facm, lu2026pixelmeanflow}, bypassing explicit stop-gradient scaling. We find that best performance is achieved with $p=1$, consistent with \cite{geng2025meanflow, geng2026improvedmeanflows, lu2026pixelmeanflow}. Since $x$-prediction with a $v$-space loss yields optimal performance \cite{li2025backtobasics}, we evaluate our Interval Denoiser under the MeanFlow loss, directly inducing the scaling coefficient $\beta = \frac{r^2}{t^4}$ (Prop. 2).  Letting $X_{\text{tar}}$ denote the regression target, this yields the loss:
\begin{equation}
\label{loss}
\mathcal{L}(\theta) = \mathbb{E}_{t, r, x_0, \epsilon} \Big[ \log \Big( \beta \big\| X_\theta - X_{\text{tar}}\big\|^2_2 + \delta \Big) \Big].
\end{equation}

\paragraph{Residual Stabilization.}
Few-step generation requires training over wide integration intervals ($t \approx 1.0, r \to 0$). Unlike pMF, which traps spatial predictions inside stop-gradients, our objective isolates the pure time derivative to ensure exact updates. Over large steps, however, pID produces extreme raw regression errors $e$ (Fig.~\ref{fig:residual_clipping}a), whereas pMF remains stable. Because logarithmic (adaptive) objectives \cite{peng2026facm, geng2026improvedmeanflows} scale gradients by $\frac{1}{e + \delta}$, these unbounded errors drive the scaling factor toward zero. This causes the gradient magnitude to vanish for pID (Fig.~\ref{fig:residual_clipping}b), suppressing the contribution of these specific samples relative to others in the batch. Consequently, the optimizer updates the network based almost entirely on easier, short-interval samples, effectively ignoring these critical large-step cases. To restore a balanced learning signal and align our stability with pMF, we apply residual clipping \cite{lu2025scm, peng2026facm}. The raw regression error $\Delta$ is computed strictly through the stop-gradient network:
\begin{equation}
    \Delta = X_{\theta^-} + \frac{t - r}{r}\text{JVP}_{X_{\theta^-}} - x_0.
\end{equation}
Clipping $\Delta$ to $[-1, 1]$ strictly bounds the variance of $e$. This prevents the adaptive gradient suppression and yields a highly stable regression target for the active network:
\begin{equation}
X_{\text{tar}} = X_{\theta^-} - \text{clip}(\Delta, -1, 1).
\end{equation}

\paragraph{Time-Sampling Curriculum.}
While residual clipping stabilizes large steps for pID, standard distributions still under-sample these intervals, limiting few-step quality for both pID and pMF. To address this, we apply a two-phase time-sampling curriculum \cite{curriculum2024flowmatching}. In contrast to $\alpha$-Flow \cite{zhang2026alphaflow}, which alters the loss objective by annealing from trajectory flow matching to MeanFlow, our curriculum maintains a fixed loss formulation and instead shifts the time-interval sampling distribution. During training, $t$ and $r$ are drawn independently with $t > r$, where adjusting the base distribution shifts the expected interval $t - r$. In Phase I, a logit-normal distribution concentrates training on short intervals ($t \approx r$, Fig.~\ref{fig:residual_clipping}c), enabling the network to accurately learn the local velocity field in complex trajectory regions. In Phase II, transitioning to a uniform distribution shifts density toward wider intervals ($t \gg r$), forcing the network to learn the generative leaps required for few-step sampling while allocating more iterations for refinement near the clean data manifold.

\section{Experiments}

\subsection{Experimental Setup}
We evaluate on ImageNet $256 \times 256$ \cite{deng2009imagenet}. For ablations, we follow the pMF-B/16 architecture, operating directly in pixel space without pre-trained autoencoders. Following \citet{geng2026improvedmeanflows}, classifier-free guidance
\cite{ho2022classifierfreediffusionguidance} is applied at training time (see Appendix~\ref{sec:cfg}). Ablation models are trained from scratch for 160 epochs, while scaled models are evaluated in main results. We report Fréchet Inception Distance (FID) \cite{heusel2017gans} and Inception Score (IS) \cite{salimans2016improved} on 50,000 samples. Implementation details are provided in Appendix~\ref{sec:impl}.

\subsection{Ablation Study}

\begin{figure}[tb]
    \centering
    \includegraphics[width=\columnwidth]{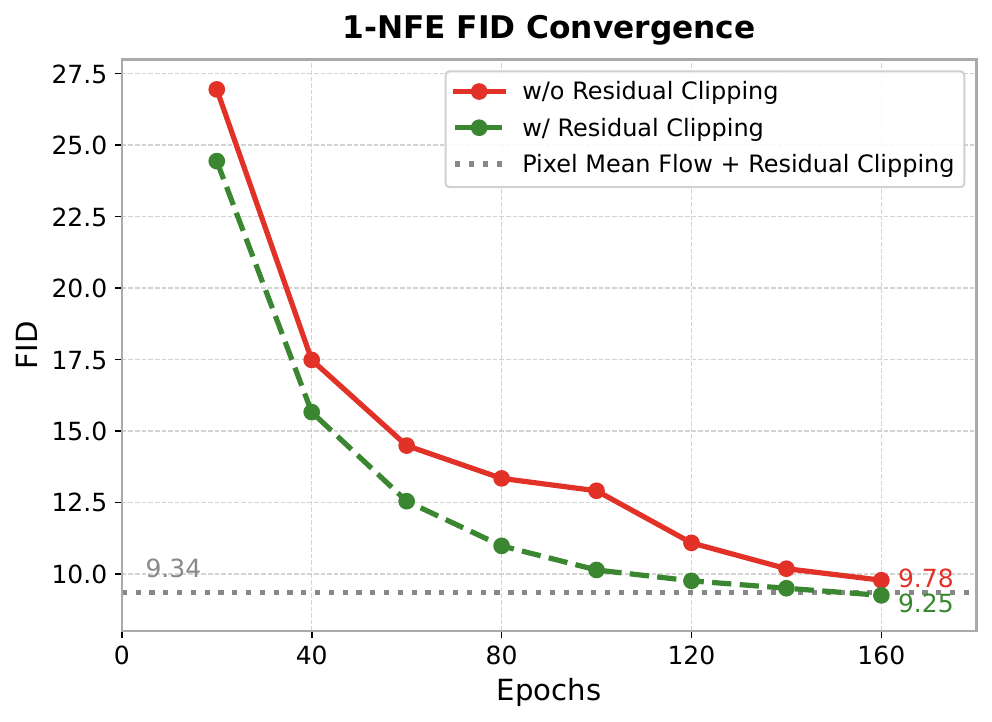} 
    \caption{Effect of residual clipping on 1-NFE training. Bounding residual variance prevents gradient suppression, accelerating convergence and improving pID 1-NFE FID from 9.78 to 9.25. With gradient stability restored, our exact formulation achieves performance (9.25) comparable to the standard pMF baseline (9.34).}
    \label{fig:residual_clipping_fid}
\end{figure}

\paragraph{Residual Clipping.}
We evaluate the empirical impact of residual clipping on 1-NFE generation quality throughout training. As shown in Fig.~\ref{fig:residual_clipping_fid}, clipping consistently accelerates convergence for pID, improving our final 1-NFE FID from 9.78 to 9.25. Meanwhile, adding residual clipping to pMF produces similar results (9.56 w/o clipping vs. 9.34 w/ clipping), aligning with our observation that unclipped pMF is inherently stable at large steps. By restoring gradient stability to pID, our exact ODE-derived parameterization achieves performance comparable to the pMF baseline.

\paragraph{Time-Sampling Curriculum.}

\begin{table}[tb]
\centering
\small 
\begin{tabular}{@{}llccrr@{}}
\toprule
\textbf{Phase I} & \multicolumn{3}{c}{\textbf{Phase II (Epoch $T_s$--End)}} & \multicolumn{2}{c}{\textbf{Metrics}} \\
\cmidrule(r){1-1} \cmidrule(lr){2-4} \cmidrule(l){5-6}
$p_1$ & $p_2$ & Mix & $T_s$ & \textbf{FID} $\downarrow$ & \textbf{IS} $\uparrow$ \\
\midrule
\multicolumn{6}{c}{\textit{Baselines (Static Sampling)}} \\
\midrule
Uniform      & Uniform      & ---   & --- & 9.26  & 178.1 \\
LN(0.0, 0.8) & LN(0.0, 0.8) & ---   & --- & 10.67 & 152.8 \\
LN(0.8, 0.8) & LN(0.8, 0.8) & ---   & --- & 9.25  & 188.0 \\
\midrule
\multicolumn{6}{c}{\textit{Curriculum Ablations}} \\
\midrule
LN(0.8, 0.8) & Uniform & 50\%  & 120 & 7.87 & 200.0 \\
LN(0.8, 0.8) & Uniform & 50\%  & 140 & 7.88 & 198.6 \\
LN(0.8, 0.8) & Uniform & 50\%  & 150 & 7.85 & 198.3 \\
\addlinespace
LN(0.8, 0.8) & Uniform & 100\% & 120 & 7.69 & 192.0 \\
LN(0.8, 0.8) & Uniform & 100\% & 140 & \textbf{7.55} & \textbf{200.7} \\
LN(0.8, 0.8) & Uniform & 100\% & 150 & 7.69 & 194.9 \\
\addlinespace
LN(0.0, 0.8) & Uniform & 100\% & 120 & 8.10 & 186.2 \\
LN(0.0, 0.8) & Uniform & 100\% & 140 & 8.01 & 185.2 \\
LN(0.0, 0.8) & Uniform & 100\% & 150 & 8.27 & 180.3 \\
\bottomrule
\end{tabular}
\caption{Ablations on time-sampling curriculum for pID. Models are trained for 160 epochs on ImageNet 256$\times$256. Phase I uses distribution $p_1$. At epoch $T_s$, Phase II introduces distribution $p_2$ with the specified mixing probability.}
\label{tab:sampler_ablation}
\end{table}

Table~\ref{tab:sampler_ablation} evaluates the impact of the two-phase time-sampling curriculum on pID. While static baselines yield around 9.25 FID, introducing a distribution shift from $\text{LN}(0.8, 0.8)$ to Uniform improves generation quality, lowering the 1-NFE FID to 7.55.

Comparing initial distributions highlights the sensitivity to the logit-normal location parameter: shifting the Phase I mean from $\mu = 0.8$ to $\mu = 0.0$ degrades the post-curriculum FID from 7.55 to 8.01. For Phase II, a full 100\% transition to the uniform distribution consistently outperforms a 50\% mix, demonstrating that the network benefits from a complete shift to wide-interval sampling. Finally, ablating the transition epoch $T_s$ reveals that switching at $T_s = 140$ achieves the optimal 7.55 FID, whereas transitioning earlier at $T_s = 120$ or later at $T_s = 150$ yields a higher FID of 7.69.

\paragraph{Sampling with 2-NFE.}
\label{app:2nfe_ablation}

\begin{figure*}[t]
    \centering
    \includegraphics[width=\textwidth]{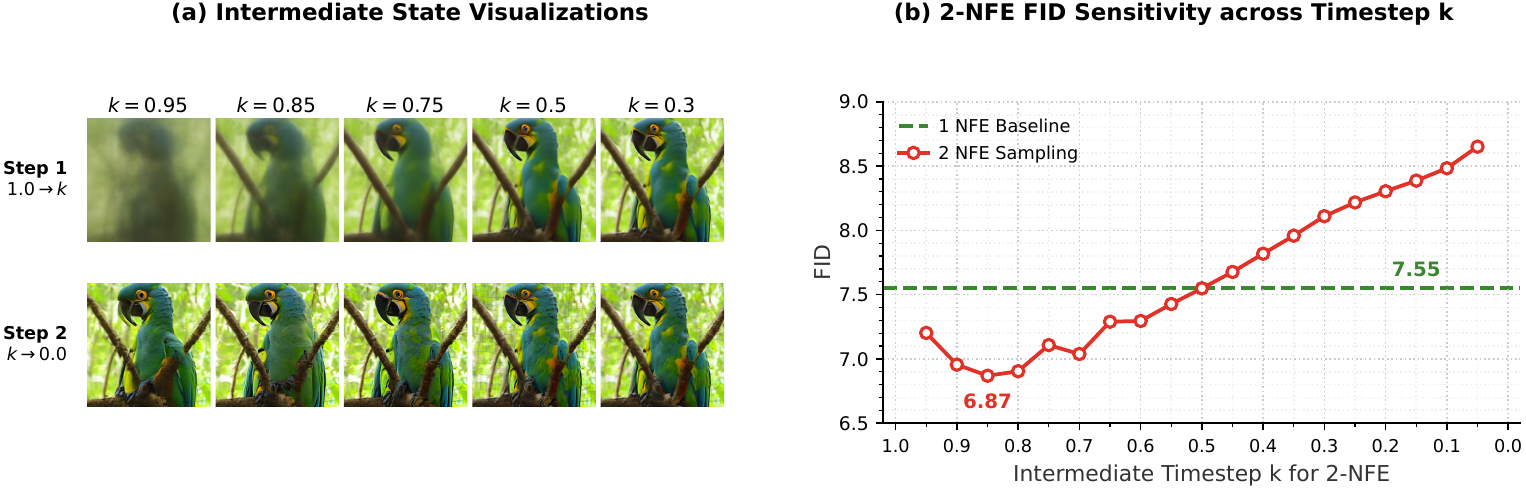}
    \caption{Analysis of 2-NFE intermediate sampling trajectory states. (a) Intermediate state visualizations after Step 1 ($1.0 \to k$, top row) and corresponding final generated images after Step 2 ($k \to 0.0$, bottom row) across different intermediate timesteps $k$. (b) 2-NFE FID sensitivity across $k$, demonstrating that 2-step sampling consistently outperforms the single-step baseline (7.55, dashed line) for $k \ge 0.5$, reaching optimal generation performance (FID 6.87) at $k = 0.85$.}
    \label{fig:2nfe_analysis}
\end{figure*}

We evaluate two-step (2-NFE) sampling across intermediate timesteps $k \in (0, 1)$. As shown in Fig.~\ref{fig:2nfe_analysis}b, performance is highly sensitive to $k$, reaching an optimal FID of 6.87 at $k = 0.85$ (vs. 7.55 for 1-NFE). Setting $k < 0.5$ degrades quality below single-step sampling. As shown in Fig.~\ref{fig:2nfe_analysis}a, a short initial step ($k=0.85$) yields a clean structural prior for Step 2 to refine, whereas smaller $k$ causes premature detail generation.

\subsection{Main Results and Comparisons}

\paragraph{Scaling Model Capacity and Training Budget.}
We evaluate scalability of our pixel Interval Denoiser (pID) by extending the training budget to 320 epochs. On the Base architecture (pID-B/16), extending training improves 1-NFE FID from 7.55 to 6.28, which further drops to 5.61 with 2-NFE sampling. Scaling to the Large configuration (pID-L/16) under the same budget yields a 1-NFE FID of 4.55 and a 2-NFE FID of 3.98. These results demonstrate strong scalability across both model capacity and training duration.

\paragraph{Comparisons on ImageNet 256$\times$256.}
Table~\ref{tab:sota_comparison} compares our model with prior generative frameworks. We explicitly differentiate pure probability flow models from those relying on auxiliary perceptual losses. As established by recent studies \cite{kynkaanniemi2023role, song2024improvedct}, training with perceptual metrics (e.g., LPIPS) causes feature leakage from ImageNet-pretrained networks. Because FID itself relies on an ImageNet-pretrained Inception-V3 classifier, this alignment artificially lowers FID scores by exploiting the metric's perceptual null space rather than improving true sample quality. Focusing strictly on direct probability distribution matching without auxiliary loss shortcuts, our pID-L/16 sets new state-of-the-art performance for pure pixel-space fast-forward models in both 1-NFE (4.55) and 2-NFE (3.98) regimes.

\begin{table}[ht!]
\centering
\small
\begin{tabular*}{\columnwidth}{@{\extracolsep{\fill}} l c c r r @{}} 
\toprule
Method & Epoch & \# Params & FID $\downarrow$ & IS $\uparrow$ \\
\midrule
\grouphead{Multi-step Pixel-space Diffusion/Flow} \\
\dimrow{JiT-L/16 \shortcite{li2025backtobasics}}{600}{459M}{2.36}{298.5} \\
\dimrow{ADM-G \shortcite{dhariwal2021diffusion}}{400}{554M}{4.59}{186.7} \\
\dimrow{RIN \shortcite{jabri2023scalable}}{480}{410M}{3.42}{182.0} \\
\dimrow{PixNerd-L/16 \shortcite{wang2025pixnerd}}{160}{458M}{2.64}{297.0} \\
\grouphead{1-NFE Latent-space Diffusion/Flow} \\
\dimrow{iCT-XL/2 \shortcite{song2024improvedct}}{---}{675M}{34.24}{---} \\
\dimrow{Shortcut-XL/2 \shortcite{frans2025shortcut}}{250}{675M}{10.60}{102.7} \\
\dimrow{MF-L/2 \shortcite{geng2025meanflow}}{240}{459M}{3.84}{250.9} \\
\dimrow{iMF-L/2 \shortcite{geng2026improvedmeanflows}}{640}{409M}{1.86}{276.6} \\
\grouphead{1-NFE Pixel-space GANs} \\ 
\dimrow{BigGAN-deep \shortcite{brock2019biggan}}{---}{56M}{6.95}{171.4} \\
\dimrow{StyleGAN-XL \shortcite{sauer2022styleganxl}}{---}{166M}{2.30}{260.1} \\
\dimrow{GigaGAN \shortcite{kang2023gigagan}}{---}{569M}{3.45}{225.5} \\
\grouphead{1-NFE Pixel-space Diffusion/Flow --- \underline{with} perceptual losses} \\
\dimrow{pMF-B/16 \shortcite{lu2026pixelmeanflow}}{320}{118M}{3.12}{---} \\
\dimrow{pMF-L/16 \shortcite{lu2026pixelmeanflow}}{320}{411M}{2.52}{---} \\
\grouphead{1-NFE Pixel-space Diffusion/Flow --- \underline{no} perceptual losses} \\
EPG-L/16 \shortcite{lei2026epg} & 560 & 540M & 8.82 & --- \\
pMF-B/16 \shortcite{lu2026pixelmeanflow} & 320 & 118M & 8.71 & --- \\
\textbf{pID-B/16 (ours)} & 320 & 118M & \textbf{6.28} & 213.1 \\
\textbf{pID-L/16 (ours)} & 320 & 411M & \textbf{4.55} & 221.9 \\
\grouphead{2-NFE Pixel-space Diffusion/Flow --- \underline{no} perceptual losses }\\
\textbf{pID-B/16 (ours)} & 320 & 118M & \textbf{5.61} & 224.5 \\
\textbf{pID-L/16 (ours)} & 320 & 411M & \textbf{3.98} & 243.0 \\
\bottomrule
\end{tabular*}
\caption{Comparison on ImageNet $256 \times 256$. FID and IS are evaluated on 50,000 generated samples. The first four groups are reference baselines that rely on latent spaces, multi-step sampling, or perceptual losses, and are not directly comparable to the pure pixel-space setting of the final two groups.}
\label{tab:sota_comparison}
\end{table}

\section{Conclusion}

We presented the \textit{Interval Denoiser}, a rigorous framework for few-step, latent-free generation. We showed that prior pixel-space methods relying on empirical algebraic substitutions trap spatial predictions inside stop-gradients, causing biased first-order updates.
To resolve this, we analytically derived the Interval Denoiser directly from the flow matching ODE, projecting intermediate trajectory states onto the low-dimensional image manifold. By algebraically isolating the pure time derivative, our formulation aligns backpropagation with true analytical gradients, enabling exact first-order optimization.

Combining our exact objective with residual clipping and a time-sampling curriculum stabilizes wide integration steps, driving superior 1-NFE performance. Trained from scratch on ImageNet $256 \times 256$, without pre-trained autoencoders or perceptual losses, our pID-L/16 model achieves an FID of 4.55 at 1-NFE and 3.98 at 2-NFE, setting new state-of-the-art among pure pixel-space fast-forward models. By establishing a mathematically grounded foundation for direct image-space regression, our framework narrows the gap with latent-space models, paving the way for efficient, tokenizer-free generative modeling.


\bibliography{references} 

\clearpage
\appendix
\setcounter{secnumdepth}{2}
\renewcommand{\thefigure}{A\arabic{figure}}
\renewcommand{\thetable}{A\arabic{table}}
\setcounter{figure}{0}
\setcounter{table}{0}

\begin{table}[t]
\centering
\small
\setlength{\tabcolsep}{4pt}
\begin{tabular}{@{}lcc@{}}
\toprule
configs & pID-B/16 & pID-L/16 \\
\midrule
epochs                  & 160$^\dagger$/\,320 & 320 \\
batch size              & \multicolumn{2}{c}{1024} \\
optimizer               & \multicolumn{2}{c}{Muon, $(\beta_1, \beta_2)\!=\!(0.9, 0.95)$} \\
learning rate           & \multicolumn{2}{c}{1e-3} \\
lr warmup               & \multicolumn{2}{c}{0 epoch} \\
weight decay, dropout   & \multicolumn{2}{c}{0.0} \\
ema half-life (Mimgs)   & \multicolumn{2}{c}{$\{1000, 2000\}$} \\
ratio of $r \neq t$     & \multicolumn{2}{c}{50\%} \\
$(t, r)$ cond           & \multicolumn{2}{c}{$t - r$} \\
cls drop                & \multicolumn{2}{c}{0.1} \\
CFG dist $\beta_{\mathrm{cfg}}$ & 1 & 2 \\
\midrule
loss                    & \multicolumn{2}{c}{$\log(\beta \|\cdot\|_2^2 + \delta)$, $\beta = r^2/t^4$} \\
$\delta$                & \multicolumn{2}{c}{0.01} \\
denom.\ clip $r_{\min}$ & \multicolumn{2}{c}{0.05} \\
residual clip           & \multicolumn{2}{c}{$[-1, 1]$} \\
phase I sampler         & \multicolumn{2}{c}{logit-normal(0.8, 0.8)} \\
phase II sampler        & \multicolumn{2}{c}{uniform} \\
transition epoch $T_s$  & \multicolumn{2}{c}{140} \\
\bottomrule
\end{tabular}
\caption{Configurations and hyper-parameters. $^\dagger$: for
ablation studies. Optimizer: Muon \cite{jordan2024muon}. Class
dropout follows \citet{goyal2017accurate}. CFG settings follow \citet{geng2026improvedmeanflows}.}
\label{tab:hparams}
\end{table}

\section{Classifier-Free Guidance}
\label{sec:cfg}
Following Improved MeanFlow \cite{geng2026improvedmeanflows}, we incorporate classifier-free guidance (CFG) \cite{ho2022classifierfreediffusionguidance} directly into training rather than at inference. By substituting our Interval Denoiser into the velocity guidance formula via the identity $u = (z_t - X)/t$, we construct the guided target:
\begin{equation}
    x_{\mathrm{cfg}} = x_0 + \left(1-\frac{1}{\omega}\right)\Big(X_{\theta^-}(z_t, t, t \mid \mathbf{c}) - X_{\theta^-}(z_t, t, t \mid \emptyset)\Big).
\end{equation}
Here, $\mathbf{c}$ and $\emptyset$ denote the conditional and unconditional classes, and $\omega$ is the guidance scale. We apply this by replacing the clean image $x_0$ with $x_{\mathrm{cfg}}$ in our regression objective. During training, both $\omega$ and the CFG interval are sampled and provided to the network as conditioning inputs.

\section{Implementation Details}
\label{sec:impl}

We use the unmodified pMF-B/16 and pMF-L/16 architectures. Detailed configurations are in Table~\ref{tab:hparams}; unspecified hyperparameters follow Pixel Mean Flow (pMF) \cite{lu2026pixelmeanflow}.

\paragraph{Denominator clipping.}
Because the coefficient $(t - r)/r$ diverges as $r \to 0$, we clip the denominator to a minimum value of $r_{\min} = 0.05$.

\paragraph{Auxiliary head.}

Following \citet{geng2026improvedmeanflows}, the network employs two jointly trained output heads: a primary head predicting the Interval Denoiser $X_\theta(z_t, r, t)$, and an auxiliary head predicting the instantaneous denoiser $x_\theta(z_t, t)$. The total training objective is the sum of the pID loss for $X_\theta$ and the flow matching loss for $x_\theta$. At inference, the auxiliary head is entirely discarded, and sampling relies only on $X_\theta$.

\paragraph{EMA.}
Following pMF, we maintain multiple Exponential Moving Average (EMA) half-lives during training and select the best for inference.

\paragraph{Longer training.}
For 320-epoch runs (pID-B/16 and pID-L/16), we use the optimal ablation settings and transition to the uniform time sampler at epoch 140.

\paragraph{Baselines.}
Our pMF-B/16 reproduction at 160 epochs (without residual clipping) yields 9.56 FID, matching the value reported by \citet{lu2026pixelmeanflow}. 

\section{Computational Budget}
\label{sec:compute}
Models are trained on a single node with 8 NVIDIA H100 GPUs. pID-B/16 requires around $3$ days for 160 epochs (576 H100-hours) and around $6$ days for 320 epochs (1,152 H100-hours). pID-L/16 takes around $18$ days for 320 epochs (3,456 H100-hours).

\section{Failed Experiments}
\label{sec:negative}

We document directions that did not improve our framework using pID-B/16 at 160 epochs with residual clipping.

\paragraph{Denominator clipping values.}
The impact of $r_{\min}$ depends on the time sampler. Under a uniform sampler, reducing $r_{\min}$ from $0.05$ to $0.01$ improves FID from 9.26 to 8.62. Conversely, under logit-normal$(0.8, 0.8)$, it degrades FID from 9.25 to 9.62. However, our two-phase curriculum eliminates this sensitivity: setting $r_{\min} = 0.01$ yields 7.53 FID, comparable to our reported 7.55 FID using $r_{\min} = 0.05$.

\paragraph{Alternative consistency techniques.}
We explore several techniques from the broader consistency model literature. These include interpolating between the regression target and the network output \cite{peng2026facm}, applying a tangent warmup to linearly scale the JVP term \cite{lu2025scm, sabour2025align}, and using alternative loss weightings \cite{kim2025meanflow}. None tangibly improved generation quality or training stability.

\section{Visualization}
\label{sec:qualitative}

\begin{figure*}[p]
\centering
\includegraphics[width=\textwidth]{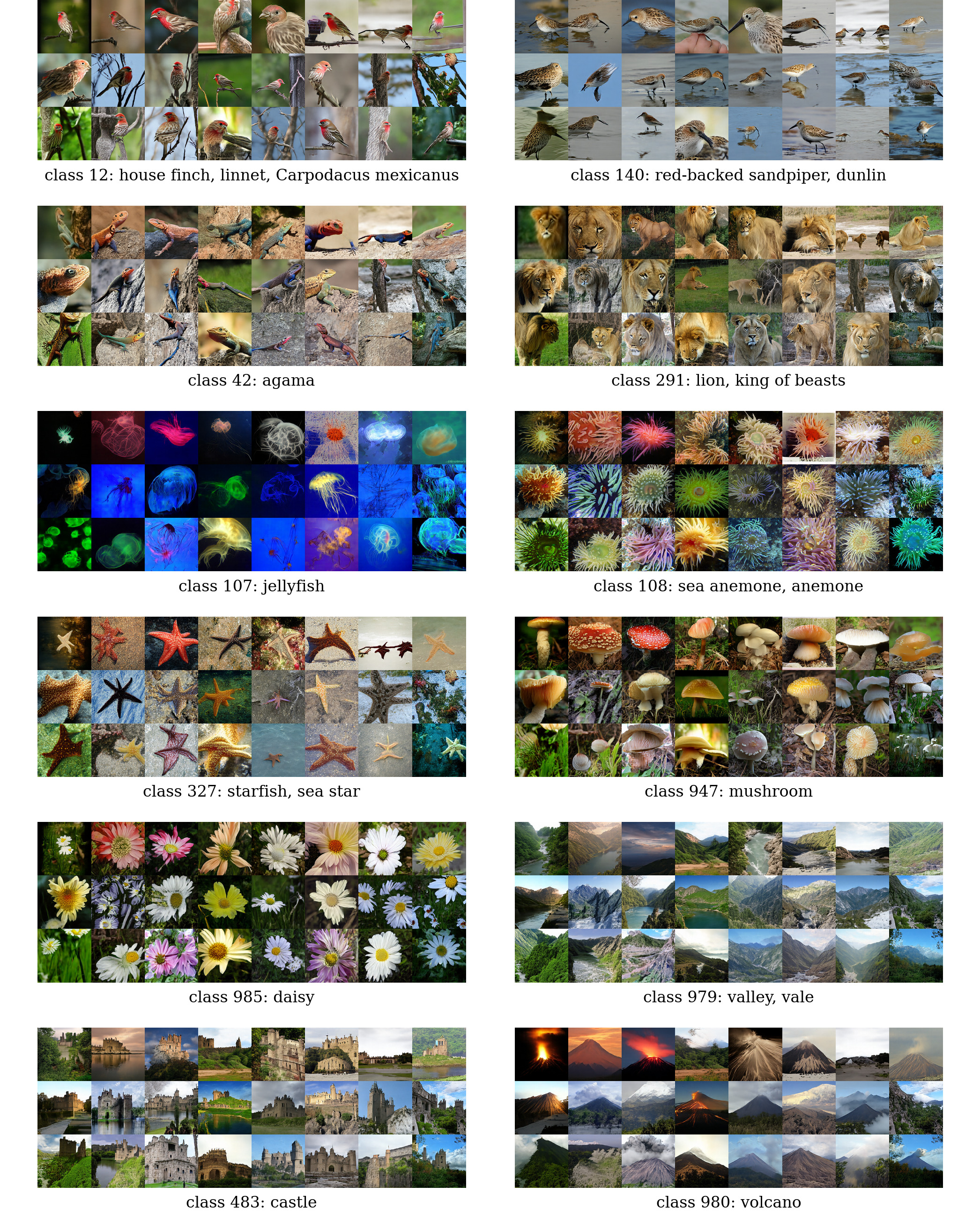}
\caption{Uncurated 1-NFE pixel class-conditional generation
samples of pID-L/16 on ImageNet $256 \times 256$.}
\label{fig:uncurated_1nfe}
\end{figure*}

\begin{figure*}[p]
\centering
\includegraphics[width=\textwidth]{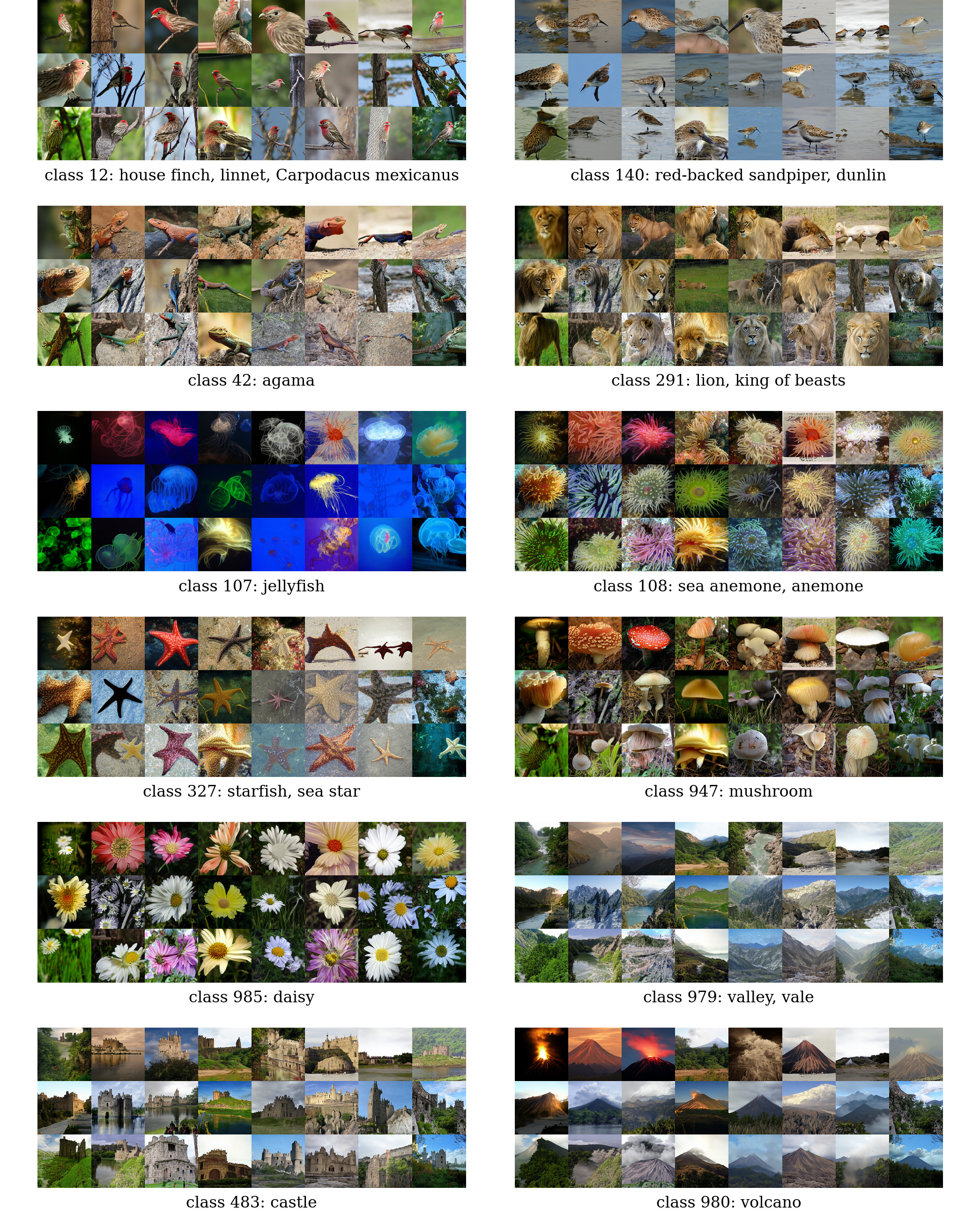}
\caption{Uncurated 2-NFE pixel class-conditional generation
samples of pID-L/16 on ImageNet $256 \times 256$.}
\label{fig:uncurated_2nfe}
\end{figure*}

Figures~\ref{fig:uncurated_1nfe} and~\ref{fig:uncurated_2nfe} provide uncurated pID-L/16 samples on ImageNet $256 \times 256$. Each block uses random seeds 1--24 in raster order. Both figures share the same initial noise, making corresponding cells directly comparable.

We use the settings from our reported evaluation: 1-NFE (FID 4.55) uses CFG scale $\omega = 7.0$ and interval $[0.1, 0.82]$; 2-NFE (FID 3.98) uses $\omega = 7.0$, interval $[0.1, 0.74]$, and intermediate timestep $k = 0.8$.

\end{document}